\documentclass[runningheads]{llncs}
\usepackage{graphicx}
\usepackage{siunitx}
\usepackage{amsmath}
\usepackage{amssymb}
\usepackage{refcount}
\usepackage{caption}
\begin{document}
\title{Instance Segmentation of Biomedical Images with an Object-aware Embedding Learned with Local Constraints\thanks{This work was supported by the Deutsche Forschungsgemeinschaft (Research Training Group 2416 MultiSenses-MultiScales).}}
\titlerunning{Instance Segmentation with an Object-aware Embedding}
%

\author{Long Chen\inst{1}\orcidID{0000-0002-5280-4727} \and
Martin Strauch\inst{1}\orcidID{0000-0001-6754-211X} \and
Dorit Merhof\inst{1}\orcidID{0000-0002-1672-2185}}
\index{Chen, Long}
\index{Strauch, Martin}
\index{Merhof, Dorit} 
\authorrunning{L. Chen et al.}
%
\institute{Institute of Imaging \& Computer Vision, RWTH Aachen University, Aachen, Germany\\
\email{\{long.chen, martin.strauch, dorit.merhof\}@lfb.rwth-aachen.de}\\
\url{https://www.lfb.rwth-aachen.de/}}

\maketitle              

\begin{abstract}
	Automatic instance segmentation is a problem that occurs in many biomedical applications. State-of-the-art approaches either perform semantic segmentation or refine object bounding boxes obtained from detection methods. Both suffer from crowded objects to varying degrees, merging adjacent objects or suppressing a valid object. In this work, we assign an embedding vector to each pixel through a deep neural network. The network is trained to output embedding vectors of similar directions for pixels from the same object, while adjacent objects are orthogonal in the embedding space, which effectively avoids the fusion of objects in a crowd. Our method yields state-of-the-art results even with a light-weighted backbone network on a cell segmentation (BBBC006 + DSB2018) and a leaf segmentation data set (CVPPP2017).The code and model weights are public available\footnote{https://github.com/looooongChen/instance\_segmentation\_with\_pixel\_embeddings/}. 
	
\end{abstract}

\section{Introduction}
\label{sec:intro}
Many biomedical applications, such as phenotyping~\cite{plantsPheno} and tracking~\cite{cellTracking}, rely on instance segmentation, which aims not only to group pixels in semantic categories but also to segment individuals from the same category. This task is challenging because objects of the same class can get crowded together without obvious boundary clues.

A prevalent class of approaches used for biomedical images is based on semantic segmentation, obtaining instances through per-pixel classification~\cite{unet,dcan}. Although this approach generates good object coverage, crowded objects are often mistakenly regarded as one connected region. DCAN~\cite{dcan} predicts the object contour explicitly to separate touching glands. However, segmentation by contours is very unreliable in many cases, since a few misclassified pixels can break a continuous boundary.

Another major class of approaches, such as Mask-RCNN~\cite{mrcnn}, refine the bounding boxes obtained from object detection methods~\cite{frcnn,ssd}. Object detection methods rely on non-maximum suppression (NMS) to remove duplicate predictions resulting from exhaustive search. This becomes problematic when bounding boxes of two objects overlap with a large ratio: one valid object will be suppressed. A finer shape representation star-convex polygons is used by~\cite{stardist} with the intention of reducing false suppression. However, it is only suitable for roundish objects~\cite{stardist,shape}. 

\begin{figure}[]
	\centering
	
	\includegraphics[width=\textwidth]{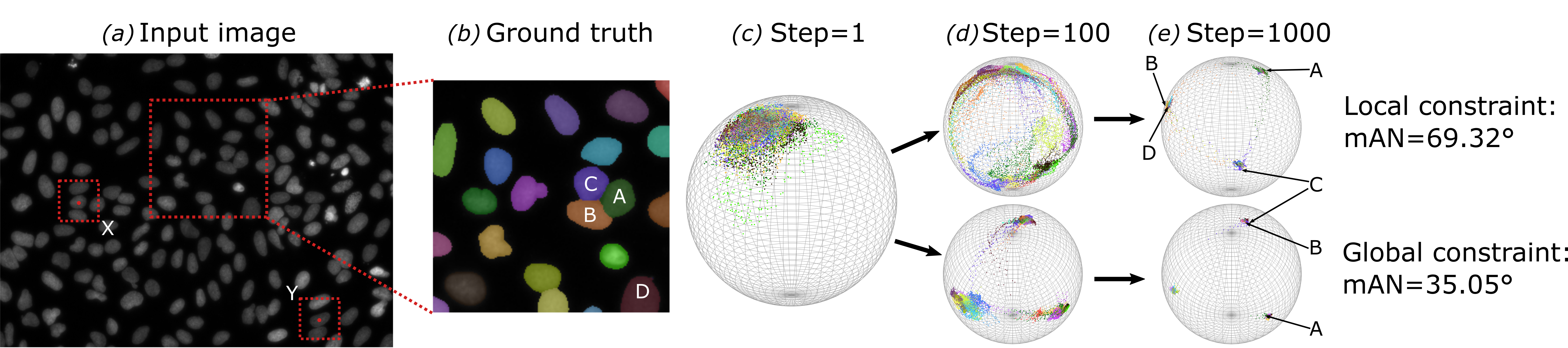}
	
	\caption{\textit{(a)} In images of repeated patterns, different pixels, such as X and Y, can have similar content in their receptive fields. \textit{(c)}-\textit{(e)} demonstrate the convergence of the embedding loss on image~(a) in a 3 dimensional space (background ignored). In \textit{(e)}, both local and global constraints form 3 clusters which are orthogonal to each other. While adjacencies A, B and C are well separated under local constraints, B and C belong to the same cluster under global constraints. The better discriminative property of local constraints is also reflected by the mean angle of neighbors ($mAN$). In addition, distant objects, such as B and D, occupying the same space is a desired property.}
	\label{fig:emb}
\end{figure}

In this work, we propose to get instances by grouping pixels based on an object-aware embedding. A deep neural network is trained to assign each pixel an embedding vector. Pixels of the same object will have similar directions in the embedding space, while spatially close objects are orthogonal to each other. Since our method performs pixel-level grouping, it is not affected by different object shape and it does not suffer from the false suppression problem. On the other hand, it avoids the fusion of adjacent objects like the semantic segmentation based methods. 

Some recent research~\cite{disAuto,disLoss,rpe} proposes the use of embedding vectors to distinguish individual objects in the driving scene and natural images. These approaches force each object to occupy a different part of the embedding space. The global constraint is actually not necessary, and could even be detrimental, for biomedical images that often contain repeated local patterns. For example, content in the receptive fields of pixel X and Y (Fig.~\ref{fig:emb}(a)) are very similar, both with one object above and one below. The network has no clear clue to assign X and Y different embeddings. Forcing them to be different is likely to hinder training. Furthermore, the global constraint is inefficient in terms of embedding space utilization. There is no risk of distant objects being merged, thus they could share the same embedding space, such as B and D in Fig.~\ref{fig:emb}. 

The main contributions of our work are as follows: (1) we propose to train the embedding mapping only constraining adjacent objects to be different, (2) a novel loss of a good geometrical explanation (adjacent instances live in orthogonal space), (3) a multi-task network head for embedding training and obtaining segmentations from embeddings, which can be applied to any backbone networks.   

Our method is compared with several strong competing approaches. It yields comparable or better results on two data sets: a combined fluorescence microscopy data set of BBBC06\footnote{\label{ds_bbbc}https://data.broadinstitute.org/bbbc/BBBC006/} and the part of DSB2018\footnote{\label{ds_dsb}https://www.kaggle.com/c/data-science-bowl-2018} used by~\cite{stardist} and the CVPPP2017\footnote{\label{ds_cvppp}https://www.plant-phenotyping.org/CVPPP2017-challenge} leaf segmentation data set.

\section{Method}
\label{sec:method}

\begin{figure}[]
	\centering
	
	\includegraphics[width=0.9\textwidth]{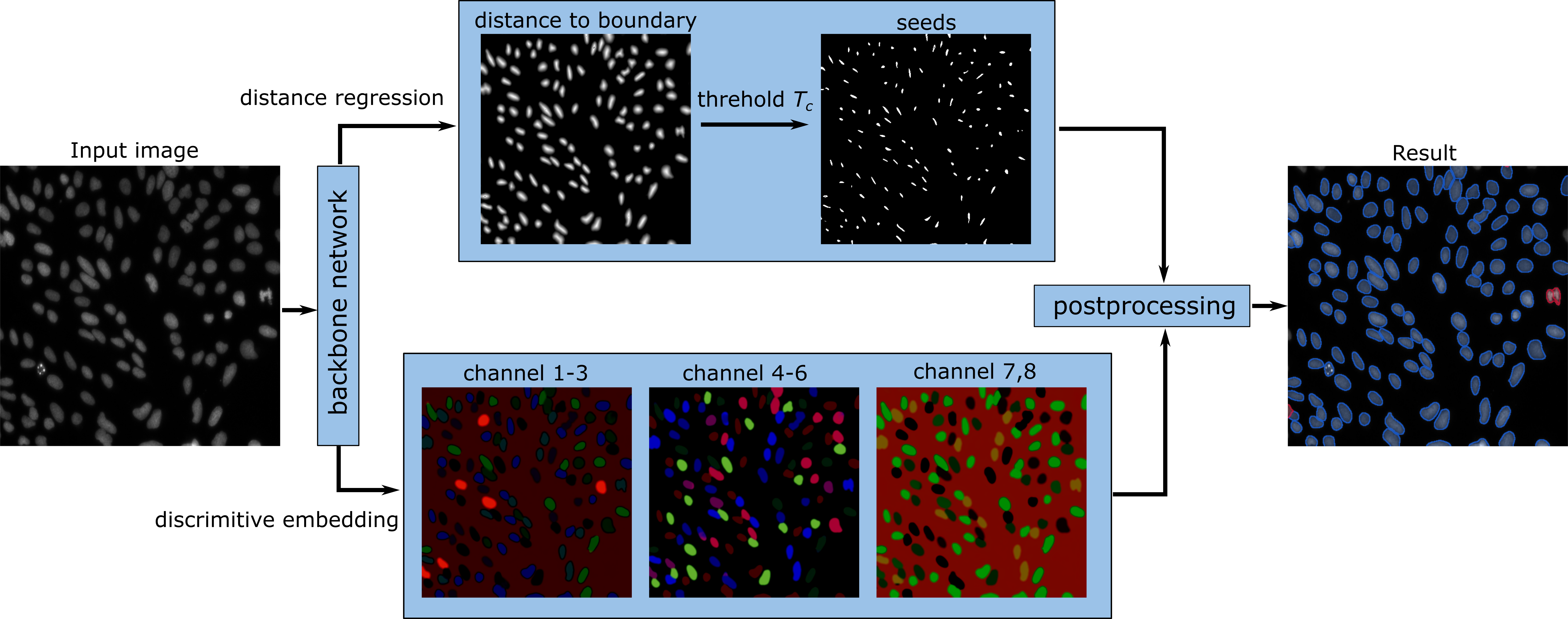}
	
	\caption{Our framework consists of two branches: the distance regression branch predicts the normalized distance from a pixel to the closest boundary, the embedding branch is responsible for mapping the feature map to the embedding space. The distance map and embedding map are combined to get segmentations (Section~\ref{sub:postprocessing}). We demonstrate the embedding as RGB images for every 3 channels.}
	\label{fig:overview}
\end{figure}

Our approach has has two output branches taking the same feature map as input: the embedding branch and the distance regression branch (Fig.~\ref{fig:overview}). Both consist of two convolutional layers. The last layer of the embedding branch uses linear activation and each filter outputs one dimension of the embedding vector.

The distance regression branch has a single layer output with relu activation. We regress the distance from an object pixel to the closest boundary pixel (normalized within each object). The distance map is used to help obtain segmentations from the embedding map, details are depicted in Section~\ref{sub:postprocessing}. 

The background is treated as a standalone object that is adjacent to all other objects. For distance regression, background pixels are set to zero. It is worth mentioning that the distance map alone provides enough cue to separate objects. But we argue that it is not optimal to obtain accurate segmentations since both object and background pixels are of small values around the boundaries, which is ambiguous and sensitive to small perturbations. In this work, the distance regression plays the role of roughly locating the objects.

\subsection{Loss function}
\label{sub:loss}
The training loss consists of two parts: $L_{reg}$ and $L_{emb}$, which supervise the learning of the distance regression branch and the embedding branch separately. We use $\lambda_1=5$ to give more emphasis on the embedding training.
\begin{equation*}
L=L_{reg}+\lambda_1 L_{emb}
\end{equation*}
We minimize the mean squared error for the distance regression, with each pixel weighted to balance the foreground and background frequency.

Intuitively, embeddings of the same object should end up at similar positions in the embedding space, while different objects should be discriminable. So naturally, the embedding loss is formulated as the sum of two terms: the consistency term $L_{con}$ and the discriminative term $L_{dis}$. 

To give a specific formula, we have to determine how "similarity" is measured. While euclidean distance is used by many works~\cite{disLoss,deepMetri}, we construct the loss with cosine distance, which decouples from the output range of different networks: $D(e_i,e_j) = 1-\frac{e_i^Te_j}{\|e_i\|_2\|e_j\|_2}$, where $e_i, e_j\in \mathbb{R} ^D$ are embeddings of pixel $i$ and $j$. The outcome of cosine distance ranges from 0 meaning exactly the same direction, to 2 meaning the opposite, with 1 indicating orthogonality.

Instead of pushing each object pair as far as possible~\cite{disLoss,rpe,deepMetri} in the embedding space (global constraint), we only push adjacent objects into each other's orthogonal space (local constraint). As shown in Fig.~\ref{fig:emb}, far away objects can occupy the same position in the embedding space, which uses the space more effectively. In the embedding map in Fig.~\ref{fig:overview}, only a few colors appears repeatedly, still ensuring that adjacent objects have different colors.  

Let's say that there are K objects within an image with $(M_1, M_2, \dots, M_K)$ pixels respectively. The loss can be written as follows:

\begin{align*}
L_{center} = &~ \frac{1}{\sum_{k=1}^{K}M_k} \sum_{k=1}^{K} \sum_{p=1}^{M_k} w_p(d_p - \widehat{d}_p)^2\\
L_{emb} = &~ L_{con}+ L_{dis} \\
= &~ \frac{1}{\sum_{k=1}^{K}M_k} \sum_{k=1}^{K} \sum_{p=1}^{M_k} w_p D(e_p, u_k)  + \frac{1}{K} \sum_{k=1}^{K} \frac{1}{|N_d(k)|}\sum_{n\in N_d(k)} |1-D(u_k, u_n) | \\
\end{align*}

\noindent , where $d_p$ and $\widehat{d}_p$ are the regression output and ground truth of pixel p, $e_p$ is the embedding of pixel $p$, $u_k$ is the mean embedding (normalized) of object $k$, $w_p$ is the factor for balancing the foreground and background frequency. $N_d(k)$ indicates the neighbors of object $k$. An object is considered as a neighbor if its shortest distance to object k is less than $d$.

\subsection{Postprocessing}
\label{sub:postprocessing}
Since objects form clusters in the embedding space, a clustering method that does not require to specify the number of clusters (e.g.\ mean shift~\cite{meanShift}) can be employed to obtain segmentations from the embedding. However, due to the time complexity of mean shift, even processing medium-size images takes tens of seconds. Since our embedding space has a good geometric explanation, we propose a simple but effective way to obtain segmentations:

\begin{enumerate}
	\item Threshold the distance map to get the central region of an object. We use $T_c=0.7$ in our experiment.  
	\item Compute the mean embedding $u_k$ of each seed region.
	\item Iteratively perform morphological dilation with a 3x3 kernel. Frontier pixels $e_i$ are included into the object, if it is not assigned to other objects and $D(e_i, u_k)$ is smaller than $T_e=0.3$. 
	\item Stop when no new pixels are included.
\end{enumerate}

Threshold $T_e$ is determined based on the fact that a pixel embedding should be closer to the ground truth object than any others in terms of angle. Thus, we set the midpoint $\ang{45}$ as the boundary, $T_e=1-cos(\ang{45})\approx0.3$   

\section{Results}

\subsection{Data sets and evaluation metrics}
In order to compare different methods, we chose two data sets that reflect typical phenomena in biomedical images:

\noindent\textbf{BBBC006+partDSB2018:} We combined the fluorescence microscopy images of cells used by~\cite{stardist} (part of DSB2018\footnotemark[\getrefnumber{ds_dsb}]) and BBBC0006\footnotemark[\getrefnumber{ds_bbbc}]. BBBC006 is a larger data set containing more densely distributed cells. We removed a small number of images without objects or with obvious labeling mistakes. The data were randomly split into 1003 training images and 230 test images. The evaluation metric was the \textit{average precision} (AP) over a range of $IoU$ (intersection over union) thresholds from 0.5 to 0.95~\footnotemark[\getrefnumber{ds_dsb}].

\noindent\textbf{CVPPP2017:} Compared to the roundish cells, the leaves in CVPPP2017 have more complex shapes and exhibit more overlap or contact. We randomly sampled 648 images for training and 162 images for testing. The results were evaluated in terms of symmetric best dice (SBD), foreground-background dice (FBD), difference in count (DiC) and absolute DiC~\cite{plantsPheno}.

\begin{figure}[]
	\centering
	
	\begin{minipage}[b]{0.19\textwidth}
		\centering
		Input 
		\includegraphics[width=\linewidth]{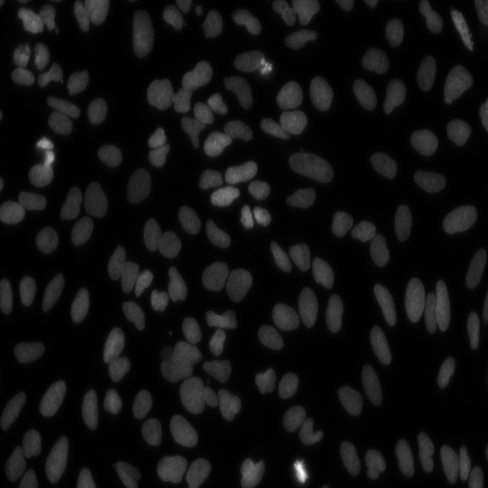} 
	\end{minipage}
	\begin{minipage}[b]{0.19\textwidth}
		\centering
		Unet 
		\includegraphics[width=\linewidth]{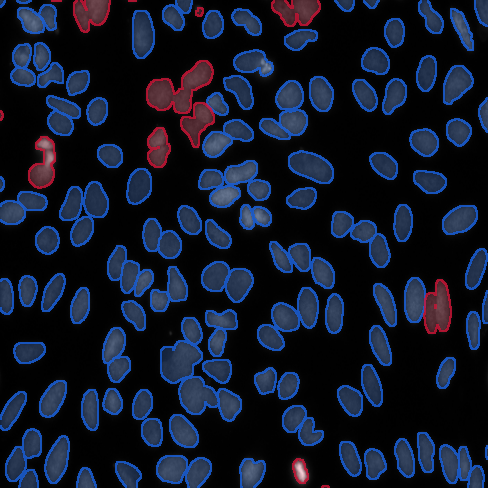} 
	\end{minipage} 
	\begin{minipage}[b]{0.19\textwidth}
		\centering
		Stardist 
		\includegraphics[width=\linewidth]{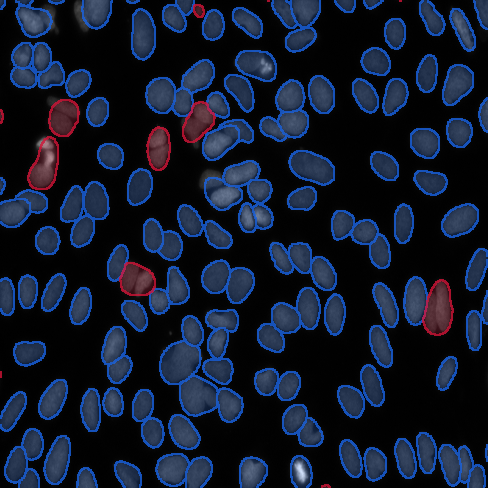} 
	\end{minipage} 
	\begin{minipage}[b]{0.19\textwidth}
		\centering
		Mask-RCNN 
		\includegraphics[width=\linewidth]{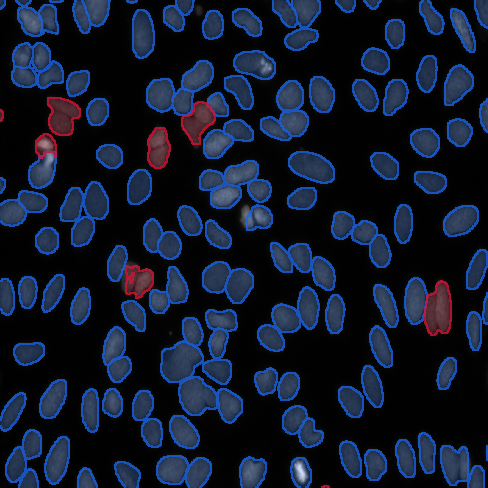} 
	\end{minipage} 
	\begin{minipage}[b]{0.19\textwidth}
		\centering
		Our method 
		\includegraphics[width=\linewidth]{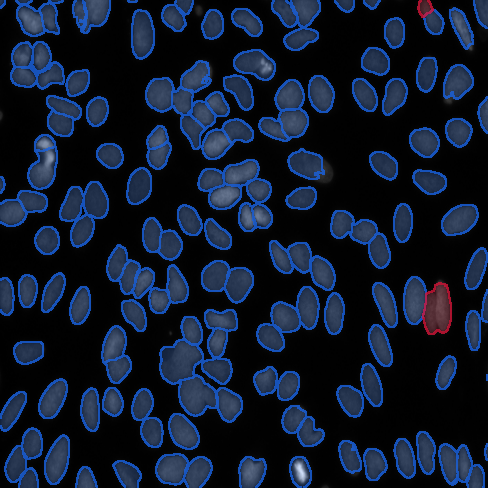} 
	\end{minipage}  
	
	\includegraphics[width=.19\textwidth]{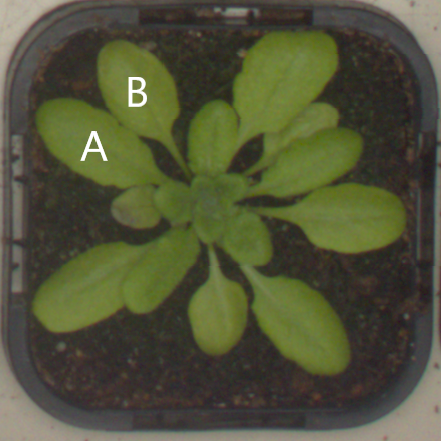}
	\includegraphics[width=.19\textwidth]{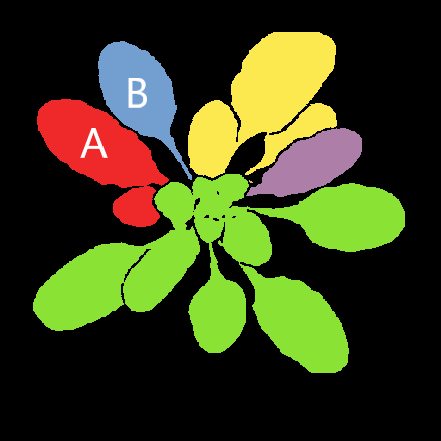}
	\includegraphics[width=.19\textwidth]{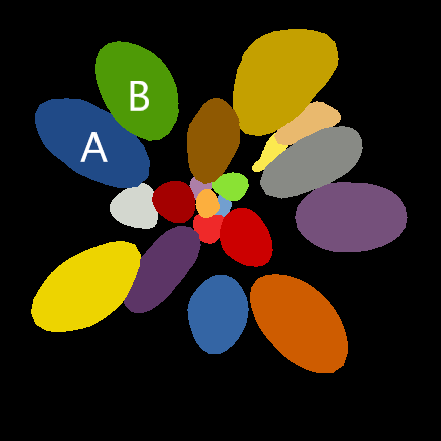}
	\includegraphics[width=.19\textwidth]{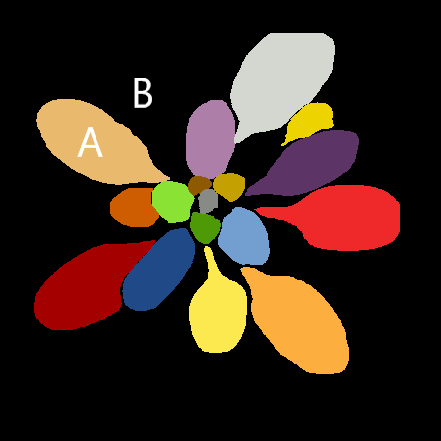}
	\includegraphics[width=.19\textwidth]{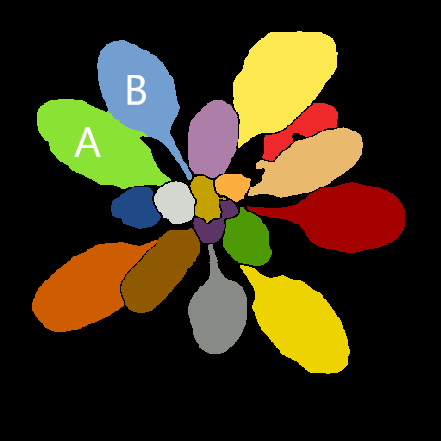}
	
	\caption{Qualitative results of the cell segmentation and leaf segmentation for four approaches. In the first row, correct matches ($IoU=0.6$) are highlighted in blue, while false positives are marked in red. The second row shows the leaf segmentation results with color-coded instances.}
	\label{fig:res}
\end{figure}

\subsection{Competing methods}
\noindent\textbf{Unet:} We employed the widely used Unet~\cite{unet} to perform 3-label segmentation (object, contour, background). Since many objects are in contact, we introduced a 2-pixel boundary to separate them.  

\noindent\textbf{Mask-RCNN:} Mask-RCNN~\cite{mrcnn} localizes objects by proposal classification and non-max suppression (NMS). Afterwards, segmentation is performed on each object bounding box. We generated 1000 proposals with anchor scales (8, 16, 32, 64, 128) for the cell data set and 50 proposals with scales (16, 32, 64, 128, 256) for the leaf data set. The NMS threshold was set to 0.9 for both data sets.

\noindent\textbf{Stardist:} Star-convex polygons are used by~\cite{stardist} as a finer shape representation. Without an explicit segmentation step, the final segmentation is obtained by combining distances from center to boundary in 32 radial directions. The final step of Stardist consists of NMS to suppress overlapping polygons.

For comparability, all methods except Mask-RCNN used a simplified U-net~\cite{stardist} (3 pooling and 3 upsampling) as the backbone network and trained from scratch. Mask-RCNN (ResNet-101~\cite{resnet} backbone) was fine-tuned on the basis of a model pretrained with the MS COCO data set\footnote{http://cocodataset.org/\#home}.

\begin{table}
	\centering
	\caption{Average precision ($AP$) for different $IoU$ thresholds on the cell data set. Different $d$ for defining neighbors (Sec.~\ref{sub:loss}) are tested (-d10, -d30 and -d100). To highlight the effect of local constraint, a 4-dimensional embedding is trained additionally.}
	\label{tab:cell}
	\begin{tabular}{*{11}{c}}
		\hline
		$IoU$ &  0.5 & 0.55 & 0.6 &  0.65 & 0.7 & 0.75 &  0.8 & 0.85 & 0.9 & $\text{mean}\; AP$ \\
		\hline
		Unet9 & .8302 & .8152 & .7994 & .7816 & .7609 & .7206 & .6216 & .4478 & .2332 & .6678\\
		Stardist &  .8178 & .8015 & .7880 & .7733 & .7552 & .7304 & .6910 & .6225 & .4749 & .7172 \\
		Mask-RCNN & .8820 & .8636 & .8492 & .8354 & \textbf{.8231} & \textbf{.8030} & \textbf{.7728} & \textbf{.7095} & .5483 & \textbf{.7874} \\
		\hline
		ours-d10-dim16 &  \textbf{.9108} &  \textbf{.8858} &  \textbf{.8611} &  \textbf{.8428} & .7936 & .7518 & .7031 & .6466 & \textbf{.5528} & .7720\\
		ours-d30-dim16 & .9039 & .8727 & .8480 & .8169 & .7776 & .7305 & .6805 & .6272 & .5311 & .7543\\ 
		ours-d100-dim16 & .9007 & .8765 & .8507 & .8212 & .7812 & .7354 & .6842 & .6256 & .5190 & .7549\\
		\hline
		ours-d10-dim4 & .9040 & .8786 & .8533 & .8130 & .7723 & .7203 & .6778 & .6254 & .5386 & .7537\\
		ours-d30-dim4 & .8925 & .8637 & .8339 & .8003 & .7525 & .7043 & .6624 & .6047 & .4878 & .7335\\ 
		ours-d100-dim4 & .6289 & .6090 & .5871 & .5567 & .5166 & .4828 & .4494 & .4082 & .3181 & .5063\\
		\hline
	\end{tabular}
\end{table}

\subsection{Results and discussion}
Th Unet had the lowest mean $AP$ in Tab.~\ref{tab:cell}. The $AP$ value decreased rapidly with increasing $IoU$ because of the false fusion of adjacent cells. Both Stardist and Mask-RCNN can handle most adjacent objects, but when a few cells form a tight roundish cluster, both methods are likely to fail. Mask-RCNN yielded the best score in the high $IoU$ range, which is the benefit of an explicit segmentation step: masks are better aligned with the object boundary. Qualitative results in Fig.~\ref{fig:res} show that our method is better at distinguishing objects that are in contact. This is also reflected by the highest $AP$ of our method for $IoU < 0.7$.

The leaf segmentation results better reflect the characteristics of each approach. As shown in Fig.~\ref{fig:res}, the Unet outlines the leaves accurately, but merges several instances into one (green and yellow). All other approaches proved to be object-aware. However, Mask-RCNN missed leaf B, because the bounding box of B is almost identical to that of A. Stardist avoids such false suppression by using a better shape representation, which comes at the expense of losing finer structures, such as the petioles. This is easy to understand, since Stardist obtains a mask by fitting a polygon based on discrete radial directions. In contrast, our method does not only avoids misses, but also produces a good contour.

\noindent\textbf{Local vs. global constraint:} To demonstrate the effect of local constraint, we tested the method with different $d$: larger $d$ treats more objects as neighbors (large enough $d$ is equivalent to the global constraint). The best result is always achieved at $d=10$, which only takes objects in contact or almost in contact as neighbors. In the case of dimension 4, the performance drop on the cell data set is especially significant at $d=100$ due to the inefficient use of embedding space. The same drop happens at $d=30$ on the leaf segmentation data set.

\begin{figure}
	\centering
	\begin{minipage}{0.6\textwidth}
		\centering
		\captionof{table}{Evaluation results on CVPPP2017 data set. See Tab.~\ref{tab:cvppp} caption for method name abbreviations.}
		\label{tab:cvppp}
		\begin{tabular}{*{5}{c}}
			\hline
			Metric &  $SBD$ & $FBD$ & $DiC$ & $|DiC|$\\
			\hline
			Unet9 & 0.5456 & 0.9045 & -3.9259 & 5.0370 \\
			Stardist & 0.8019 & 0.9327 & 1.9506 & 2.6543 \\
			Mask-RCNN & 0.7972 & 0.9060 & -0.1543 & 1.080 \\
			\hline
			ours-dist10-dim16 & \textbf{0.8307} & \textbf{0.9417} & \textbf{-0.1790} & \textbf{0.7346}\\
			ours-dist30-dim16 & 0.8159 & 0.9303 & -0.2160 & 0.7593\\
			ours-dist100-dim16 & 0.8101 & 0.9312 & -0.2593 & 0.9259\\
			\hline
			ours-d10-dim4 & 0.8005 & 0.9377 & -0.6605 & 1.0185 \\
			ours-d30-dim4 & 0.7163 & 0.3338 & -0.6358 & 0.9444 \\
			ours-d100-dim4 & 0.7095 & 0.3495 & -0.5432 & 1.0741 \\
			\hline
		\end{tabular}
	\end{minipage}
	\hspace{1em}
	\begin{minipage}{0.3\textwidth}
		\centering
		\includegraphics[width=\linewidth]{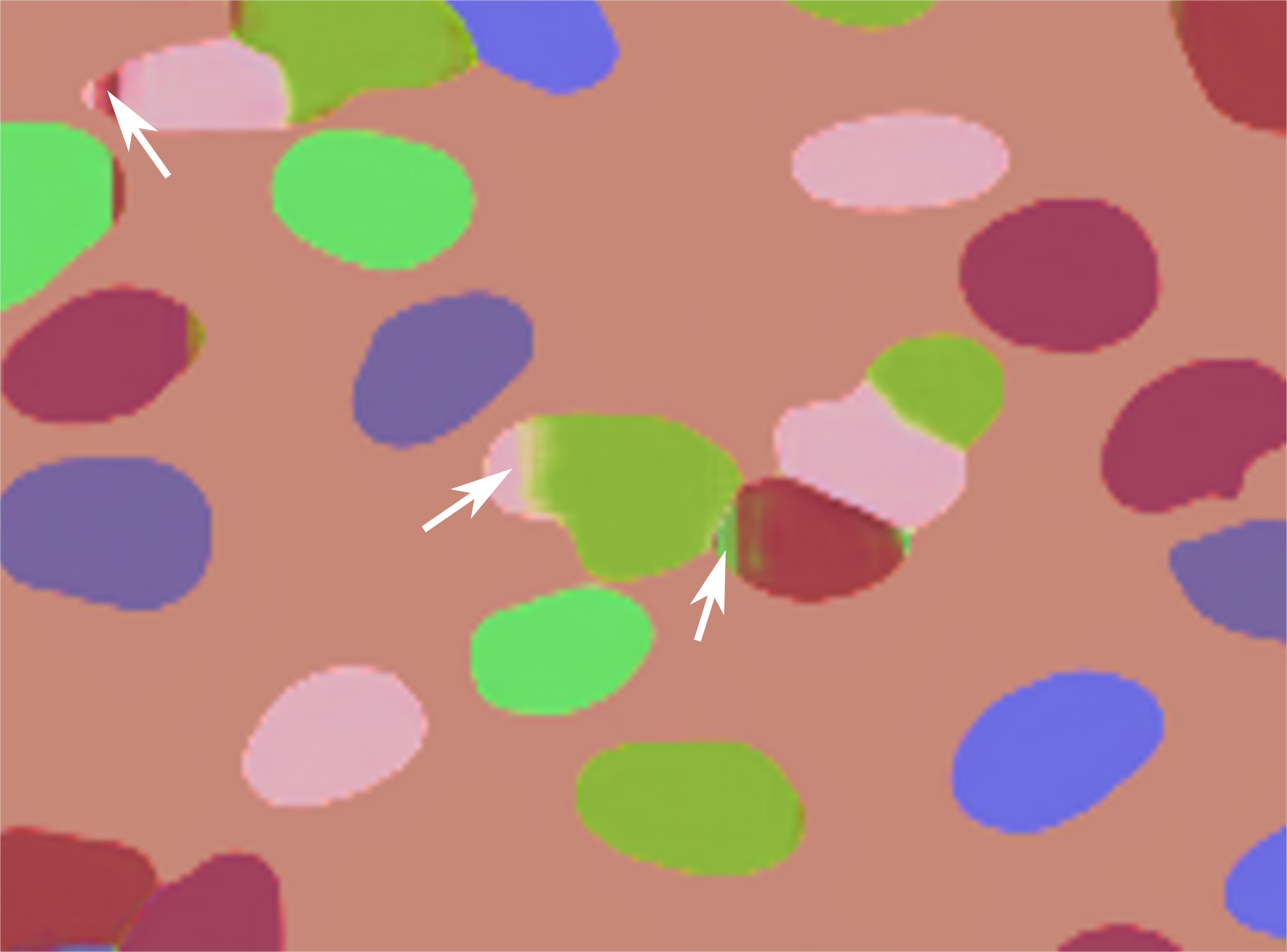} 
		\caption{Embeddings within the same object are not completely consistent (white arrows) in some cases.}
		\label{fig:artifacts}
	\end{minipage}
\end{figure}

\noindent\textbf{Incomplete object mask:} Inconsistent embeddings within an object (Fig.~\ref{fig:artifacts}) sometimes occurs near the boundary, leading to incomplete segmentations. This is why our method performs not as good as Mask-RCNN in high $IoU$ range. The reason of the inconsistence deserves further study.   


\section{Conclusion and outlook}
\label{sec:conclusion}
Our proposed approach can not only outline objects accurately, but also is free from false object suppression and object fusion. The local constraint (orthogonality of neighboring objects) makes full use of the embedding space and gives a good geometric interpretation. Our method is especially attractive for images containing a large number of objects that are repeated and in contact and yields state-of-the-art results even with a light-weighted backbone network. 

Since our approach generates embeddings that live in orthogonal spaces, if this space can be aligned with the standard space by rotating, segmentations can directly obtained from embeddings. An alternative approach to bypass postprocessing would be to add sparsity constraints on the embedding vector during training. We will test the feasibility of these two methods in the future.
%
%
%

\begin{thebibliography}{8}
	
	\bibitem{plantsPheno}
	Scharr, H., Minervini, M., French, A.P., Klukas, C., Kramer, D.M., Liu, X., Luengo, I., Pape, J., Polder, G., Vukadinovic, D., Yin, X., Tsaftaris, S.A.: Leaf segmentation in plant phenotyping: a collation study. Machine Vision and Applications, \textbf{27}(4), 585--606 (2016)
	
	\bibitem{cellTracking}
	Ulman, V., Maška, M., Magnusson, K.E.G., Ronneberger, O., Haubold, C., Harder, N., Matula, P., Matula, P., Svoboda, D., Radojevic, M., Smal, I., Rohr, K., Jald{\'e}n, J., Blau, H.M., Dzyubachyk, O., Lelieveldt, B., Xiao, P., Li Y., Cho, S.Y., Dufour, A., Olivo-Marin, J.C., Reyes-Aldasoro, C.C., Solis-Lemus, J.A., Bensch, R., Brox, T., Stegmaier, J., Mikut, R., Wolf, S., Hamprecht, F.A., Esteves, T., Quelhas, P., Demirel,  {\"O}., Malmstr{\"o}m,  L., Jug, F., Tomanc{\'a}k, P., Meijering, E., Mu{\~{n}}oz-Barrutia, A., Kozubek, M., Ortiz-de-Solor, C.: An Objective Comparison of Cell-tracking Algorithms. Nature Methods, \textbf{14}, 1141-1152 (2017)
	
	\bibitem{unet}
	Ronneberger, O., Fischer, P., Brox, T.: U-Net: Convolutional Networks for Biomedical Image Segmentation. In: 2015 MICCAI, 234--241
	
	
	\bibitem{dcan}
	Chen, H., Qi, X., Yu L., Dou, Q., Qin, J., Heng, P.A.: DCAN: Deep Contour-Aware Networks for Accurate Gland Segmentation. In: 2016 CVPR, 2487-2496
	
	\bibitem{frcnn}
	Ren, S., He, K, Girshick, R., Sun, J,: Faster R-CNN: Towards Real-time Object Detection with Region Proposal Networks. In: 28th NIPS, 91-99
	
	\bibitem{ssd}
	Liu, W., Anguelov, D., Erhan, D., Szegedy, C., Reed, S.E., Fu, C.Y., Berg, A.C.: SSD: Single Shot MultiBox Detector. In: 2016 ECCV, 21-37
	
	\bibitem{mrcnn}
	He, K., Gkioxari, G., Dollár, P., Girshick, R.: Mask R-CNN. In: 2017 ICCV, 2980-2988
	
	\bibitem{stardist}
	Schmidt, U., Weigert, M., Broaddus, C., Myers, E.W.: Cell Detection with Star-Convex Polygons. In: 2018 MICCAI, 265--273
	
	\bibitem{shape}
	Jetley, S., Sapienza, M., Golodetz, S., Torr, P.H.: Straight to shapes: Real-time detection of encoded shapes. In: 2017 CVPR, 4207-4216
	
	\bibitem{disLoss}
	De Brabandere, B., Neven, D., Van Gool, L.: Semantic Instance Segmentation with a Discriminative Loss Function. In: 2017 CoRR
	
	\bibitem{deepMetri}
	Fathi, A., Wojna, Z., Rathod, V., Wang, P., Song H.O., Guadarrama, S., Murphy, K.P.: Semantic Instance Segmentation via Deep Metric Learning. In: 2017 CoRR
	
	\bibitem{disAuto}
	De Brabandere, B., Neven, D., Van Gool, L.: Semantic Instance Segmentation for Autonomous Driving. In: 2017 CVPR Workshop, 478--480
	
	\bibitem{rpe}
	Kong, S., Fowlkes, C.C.: Recurrent Pixel Embedding for Instance Grouping. In: 2018 CVPR, 9018--9028
	
	\bibitem{meanShift}
	Comaniciu, D., Meer, P.: Mean Shift: A Robust Approach Toward Feature Space Analysis. IEEE Transactions on Pattern Analysis and Machine Intelligence, \textbf{24}(5), 603--619 (2002)
	
	\bibitem{resnet}
	He, K., Zhang, X., Ren, S., Sun, J.: Deep Residual Learning for Image Recognition. In: 2016 CVPR, 770-778
	
\end{thebibliography}
%

\end{document}